\PassOptionsToPackage{dvipsnames,table}{xcolor}
\documentclass[bookmarks=true]{article}

\usepackage[preprint]{corl_2025} 
\usepackage{times}

\usepackage{subcaption}

\usepackage{threeparttable}
\usepackage{tabularx}
\usepackage{amsmath}
\usepackage{acronym}
\usepackage{todonotes}
\usepackage{threeparttable}
\usepackage{color}
\usepackage{mathtools}
\usepackage{booktabs}
\usepackage{multirow}
\usepackage{bm}
\usepackage{soul}
\usepackage{graphicx,epstopdf}
\usepackage[linesnumbered,boxed,ruled,commentsnumbered]{algorithm2e}

\usepackage{amsmath} 
\usepackage{amssymb} 
\usepackage{algpseudocode} 
\usepackage{algorithmicx} 
\usepackage{graphicx} 
\usepackage{amsfonts} 
\usepackage{caption} 

\newif\ifuseboldmathops
\newif\ifuseittextabbrevs
\useboldmathopstrue   

\ifuseittextabbrevs

\else

\fi

\ifuseboldmathops

\else

\fi

\ifuseboldmathops

\else

\fi

\ifuseboldmathops

\else

\fi

\ifuseboldmathops

\else

\fi

\newcommand{\calF}{\mathcal{F}}

\renewcommand{\vec}[1]{\bm{#1}}

\acrodef{mdp}[MDP]{Markov decision process}
\acrodef{dfa}[DFA]{deterministic finite-state automaton}
\acrodef{ltl}[LTL]{linear temporal logic}
\acrodef{ltlf}[LTL$(\calF)$]{quantitative linear temporal logic}
\acrodef{ag}[AG]{Assume-Guarantee}
\acrodef{ssp}[SSP]{Stochastic Shortest Path}
\acrodef{mcmc}[mcmc]{Monte Carlo Markov chain}

\acrodef{ltl}[LTL]{linear temporal logic formula}
\acrodef{mdp}[MDP]{Markov decision process}
\acrodef{smdp}[Semi-MDP]{Semi-Markov decision process}

\acrodef{scltl}[scLTL]{syntactically co-safe LTL}

\title{Robust and Safe Multi-Agent Reinforcement Learning with Communication for Autonomous Vehicles: From Simulation to Hardware}

\author{
\textbf{Keshawn Smith} \\
Department of ECE \\
University of Connecticut\\
\texttt{keshawn.smith@uconn.edu}
\and
\textbf{Zhili Zhang} \\
School of Computing \\
University of Connecticut\\
\texttt{zhili.zhang@uconn.edu}
\and
\textbf{H M Sabbir Ahmad} \\
Department of ECE \\
Boston University\\
\texttt{sabbir92@bu.edu}
\and
\textbf{Ehsan Sabouni} \\
Department of ECE \\
Boston University\\
\texttt{esabouni@bu.edu}
\and
\textbf{Mainak Mondal} \\
School of Computing \\
University of Connecticut\\
\texttt{mainak.mondal@uconn.edu}
\and
\textbf{Song Han} \\
School of Computing \\
University of Connecticut\\
\texttt{song.han@uconn.edu}
\and
\textbf{Wenchao Li} \\
Department of ECE \\
Boston University\\
\texttt{wenchao@bu.edu}
\and
\textbf{Fei Miao} \\
School of Computing \\
University of Connecticut\\
\texttt{fei.miao@uconn.edu}
}

\begin{document}
\maketitle
\thispagestyle{empty}
\pagestyle{empty}

\begin{abstract}
    Deep multi-agent reinforcement learning (MARL) has demonstrated strong performance in simulation for connected autonomous driving systems. However, most prior work assumes instantaneous and perfectly synchronized inter-agent communication, limiting reliable transfer from simulation to hardware, where vehicle-to-vehicle (V2V) communication is inherently delayed and asynchronous. In practical multi-vehicle environments, shared information is subject to measurable latency and transmission variability, yet these effects are rarely incorporated into MARL training or validated on hardware. To address this gap, we propose \textbf{RSR-RSMARL}, a Robust and Safe MARL framework designed for Real-Sim-Real policy transfer under realistic communication constraints. We first measure communication latency in our hardware V2V system and use these statistics to parameterize delay-aware state sharing during simulation training. The policy is trained under both fixed and time-varying delay models to improve robustness to wireless latency. A modular Control Barrier Function (CBF)-based safety shield provides formal safety guarantees during execution. We validate the proposed framework in CARLA simulation and on physical 1/10th-scale connected autonomous vehicles (CAVs) with onboard V2V communication. Experimental results demonstrate improved coordination stability and safety under realistic latency conditions, highlighting the importance of hardware-grounded communication modeling for scalable and reliable real-to-sim-to-real MARL deployment.
\end{abstract}

\section{Introduction}
The U.S. Department of Transportation (USDOT) has recently outlined a national strategy to deploy Vehicle-to-Everything (V2X) technologies on U.S. roadways, with the goal of reducing traffic-related fatalities and improving overall transportation safety~\cite{USDOT}. As the potential of Connected and Autonomous Vehicles (CAVs) continues to attract interest, a growing body of research has explored their applications and societal benefits~\cite{zhang2023spatial, Hyldmar_2019, Miller_2020}. Multi-Agent Reinforcement Learning (MARL) has emerged as a promising paradigm for decision-making in autonomous driving, demonstrating the ability to learn cooperative and adaptive strategies in dynamic traffic environments~\cite{zhang2023spatial, Zhang_2024, Han_2022, marl_ieee_survey, han2022behavior}. By leveraging inter-agent communication, MARL-based CAV coordination strategies can enhance road safety, mitigate congestion, and improve efficiency. Despite these advances, no open-source testbed currently supports the full development and evaluation of MARL methods for CAVs. Existing testbeds are often closed-source, incomplete, or lack the necessary components to validate end-to-end multi-agent frameworks~\cite{Testbed_survey24, Hyldmar_2019, Shao2019, Tang2024}. This gap significantly limits progress toward validating robust and safe MARL algorithms in realistic CAV contexts.

Deep reinforcement learning has also been widely applied in robotics~\cite{Tang2024}, yet the sim-to-real gap remains a fundamental challenge. Policies trained in simulation often degrade in performance when deployed in real-world systems due to unmodeled noise, uncertainties, and diverse operating scenarios. Domain randomization and related techniques~\cite{MARL_hardware_ICRA25, MARL_race_CoRL23} can mitigate this discrepancy but at the cost of substantially higher computational requirements. For multi-agent CAV systems, these challenges are compounded by the critical importance of safety: unsafe actions may cause irreversible failures in densely populated environments. Robustness is further challenged by state estimation errors, communication delays, and model uncertainties. Therefore, frameworks for MARL in CAVs must incorporate safety guarantees not only during training but also throughout deployment in real-world environments.

In this work, we present a robust \textbf{Real-Sim-Real Multi-Agent Reinforcement Learning (RSR-MARL)} framework that systematically bridges high-level cooperative decision-making with safety-critical vehicle control. Rather than relying solely on domain randomization or post-hoc safety corrections, our approach integrates structured robustness directly into the training process. High-level decision-making tasks such as lane-keeping and lane-changing are handled by robust MARL policies~\cite{han2022solution, liang2022efficient}, while a CBF-based \textit{Safety Shield} enforces formal safety constraints by dynamically filtering unsafe actions during both training and deployment. This modular architecture is controller-agnostic and integrates seamlessly with different low-level controllers (e.g., PID and MPC) without altering the learning formulation. Furthermore, we explicitly model inter-agent communication delays during training, exposing policies to realistic observation latency and enabling inherent robustness to delayed information exchange. Together, delay-aware learning, formal safety enforcement, and controller-flexible design yield policies that maintain strong safety-performance trade-offs and support zero-shot transfer to real-world autonomous platforms.



The key contributions of this work are summarized as follows. 
\textbf{(1)} We propose \textbf{RSR-RSMARL}, a robust and safety-critical MARL framework with inter-agent communication designed for \textbf{Real-Sim-Real} policy transfer. The framework integrates delay-aware training and a CBF-based Safety Shield to explicitly address communication latency, model uncertainties, and state estimation errors while maintaining formal safety guarantees. 

\textbf{(2)} We develop a communication-aware latency modeling and training strategy that explicitly incorporates both fixed and time-varying delays into the MARL learning process. Unlike prior work assuming ideal communication, we inject stochastic latency during training to mimic realistic wireless delays, enabling delay-robust coordination before deployment.

\textbf{(3)} We conduct extensive validation in CARLA and on physical 1/10th-scale connected autonomous vehicle platforms, performing structured ablation studies under increasing communication delay and demonstrating zero-shot transfer of simulation-trained policies for safe real-world operation.

Our results demonstrate that the proposed \textbf{RSR-RSMARL} pipeline provides a structured and reliable pathway for transferring MARL-based cooperative decision-making from simulation to physical systems while preserving strong safety-performance trade-offs. This work represents a step toward the practical deployment of robust, communication-aware multi-agent CAV autonomy in real-world environments.

\section{Related Work}
\label{sec:related}
In this section, we review the existing literature in this area, highlighting its limitations to establish the motivation for our proposed approach.

\textbf{Deep Reinforcement Learning in Robotics} Training RL or MARL policies in simulation ensures safety and efficiency by mitigating risks to hardware and its surroundings. While imitation learning is commonly used for policy transfer \cite{Torne2024, imitation_2}, its reliance on real-world data often incurs significant costs. Our approach focuses on simulator-based training to reduce this dependency while maintaining robust real-world performance. Addressing the challenges of sim-to-real transfer, prior studies have introduced techniques such as domain randomization, state normalization, and noise injection to bridge the sim-to-real gap \cite{Zhao2020, jiang2024transic, pmlr-v155-sandha21a}. Building on these advancements, our proposed RSR-RSMARL framework aligns simulator and real-world environments by designing state and action spaces based on hardware capabilities, enabling efficient policy deployment for execution.

\textbf{Multi-Agent Systems and CAV Testbeds} Existing multi-agent system and CAV vehicular testbeds \cite{Testbed_survey24, Hyldmar_2019, Shao2019, Tang2024, Blumenkamp2024} address diverse research areas such as planning and control, computer vision, collective behavior, autonomous racing, and human-computer interaction. For instance, Blumenkamp et al.~\cite{Blumenkamp2024} introduced the Cambridge RoboMaster platform, which leverages customized DJI RoboMaster S1 robots with a tightly integrated hardware, control, and simulation stack. While effective, their approach requires a bespoke simulation environment tailored specifically to their robot platform in order to train MARL policies, limiting portability to other systems. Moreover, their framework does not incorporate explicit safety filtering during policy execution, whereas our work introduces a CBF-based Safety Shield (via QP formulations) and CBF/CLF-constrained MPC backend to enforce real-time safety guarantees. By contrast, our proposed testbed provides a fully open-source and extensible framework, supports Real-Sim-Real transfer without reliance on robot-specific simulators, and integrates robust MARL with modular safety mechanisms to ensure reliable multi-agent autonomy across diverse scenarios.

\textbf{Safe and Robust RL and MARL} Safety has become a critical focus in RL and MARL, with prior work exploring safety shields, barrier functions, and CBF-PID controllers \cite{brunke2022safe, MARLShield_21, cai2021safe, zhang2023spatial, wang2023multi, han2022behavior}, as well as robust RL methods for uncertain observations \cite{liang2022efficient, han2022solution, he2023robust}. However, these approaches often overlook the combined challenges of communication latency, sensing uncertainty, and explicit safety guarantees in multi-agent deployment. Our work advances this area by introducing \textbf{RSR-RSMARL}, a Real-Sim-Real framework that aligns simulator states and actions with hardware, incorporates V2V delays during training, and enforces safety through a CBF-based Safety Shield with pluggable PID or MPC controllers. Experiments in both CARLA and on 1/10th-scale vehicles demonstrate how this integration supports safer and more generalizable MARL-based coordination compared to existing approaches.
\section{Approach}
We introduce \textbf{RSR\text{-}RSMARL}, a Real\text{-}Sim\text{-}Real robust and safe multi-agent reinforcement learning framework centered on communication-aware policy design for connected autonomous systems. Unlike prior approaches that assume idealized information exchange, RSR\text{-}RSMARL explicitly incorporates \emph{measured real-world communication latency} into both the MARL formulation and simulation training process. By grounding training in experimentally observed delay statistics, the framework enables zero-shot transfer of learned policies to physical testbeds operating under realistic, delayed, and asynchronous information sharing conditions.

\begin{figure*}[!t]
    \centering
    {\includegraphics[width=1.0\linewidth]{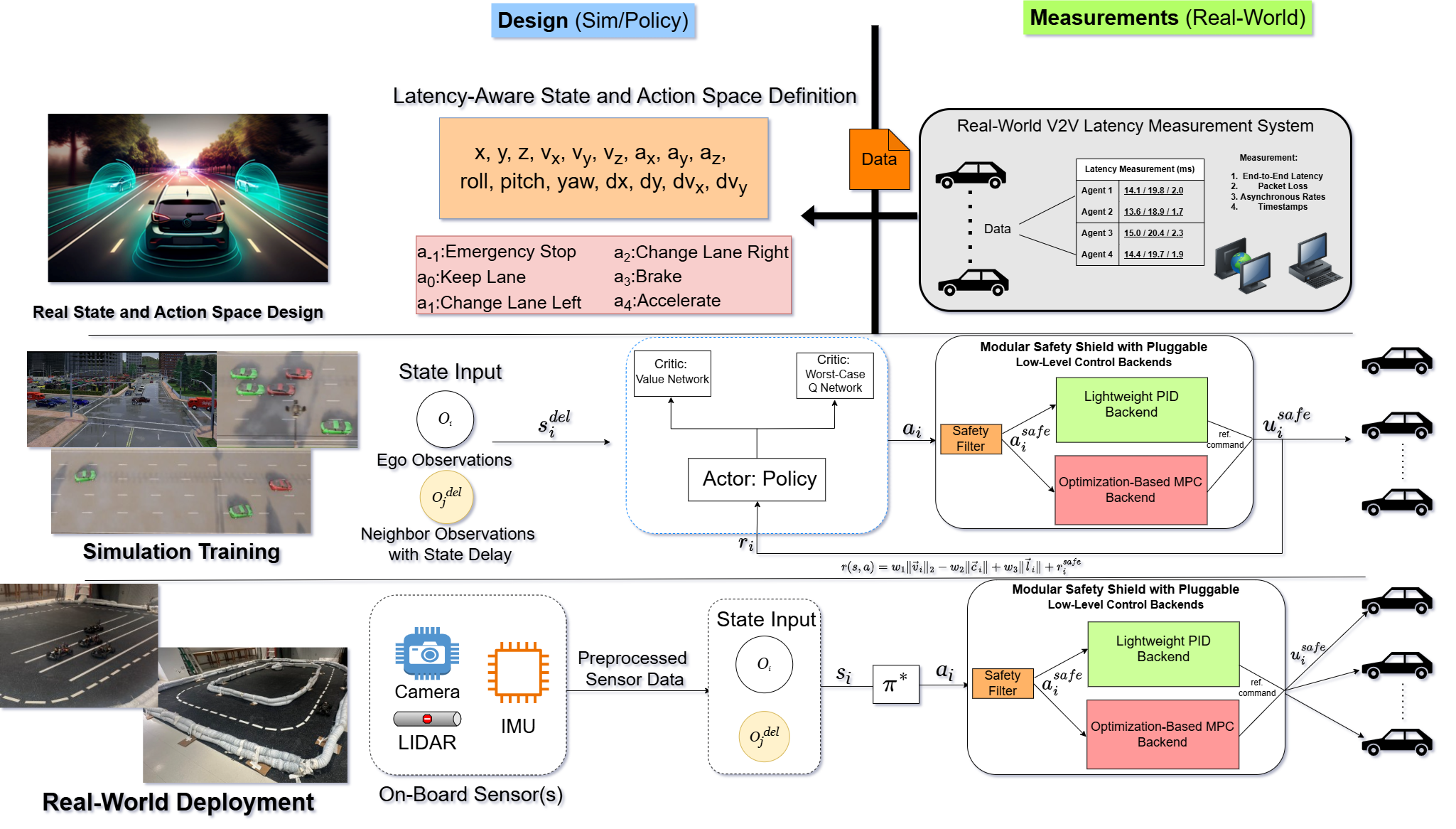}}
    \caption{Communication-aware RSR-RSMARL framework. Measured real-world V2V latency informs delay-aware state design and simulation training under fixed and time-varying delay models within a decentralized CTDE MARL architecture. A modular CBF-based Safety Shield with pluggable PID or MPC backends enforces safety during both training and deployment. Trained policies are transferred zero-shot to hardware, where delayed neighbor information and conservative safety filtering ensure robust operation under realistic communication constraints.}
    \label{fig:rsr_rsmarl}
\end{figure*}

The architecture (Fig.~\ref{fig:rsr_rsmarl}) integrates: (i) a communication-aware state and action design that models delayed neighbor information based on measured V2V latency while aligning actions with hardware constraints; (ii) latency-informed robust MARL training under fixed and time-varying delay conditions with a CBF-based safety shield; and (iii) zero-shot deployment on hardware platforms with onboard sensing and V2V communication to validate performance under real-world latency.

\subsection{MARL Formulation with Real-to-Sim Communication-Aware Design}
The design of a robust and safe MARL formulation begins with a grounded understanding of real-world operating constraints to ensure practical deployment across the simulation-to-reality gap. In this work, the primary real-to-sim design focus is \textbf{communication-aware state and action modeling}, driven by measured V2V communication latency observed on our hardware platform. Rather than assuming instantaneous or perfectly synchronized information exchange, we explicitly incorporate real-world communication delay characteristics into both the MARL problem formulation and the training process. By aligning inter-agent information sharing, state representations, and action execution with measured latency profiles, the framework avoids unrealistic communication assumptions commonly present in purely simulated approaches.

In the multi-agent setting, each connected autonomous vehicle (CAV), referred to as the ego agent $i$, receives its local observation $o_i$ and shared observations from neighboring vehicles $\mathcal{N}_i$ via V2V protocols. Crucially, shared observations are not treated as instantaneous. Instead, they are modeled as delayed and potentially asynchronous, reflecting measured fixed and time-varying latency statistics from the physical communication system. The shared information includes complementary perception features such as obstacle presence and lane boundary detections that may not be directly observable by the ego vehicle. These measurements are derived from onboard sensing modules including LiDAR and camera-based lane detection models such as UFLD~\cite{qin2020ultra}, among others~\cite{Li2022}. By explicitly modeling delayed shared states, the MARL formulation captures the temporal misalignment and uncertainty inherent in real-world connected autonomy. The MARL policy execution flow is illustrated in Figure~\ref{fig:MARVEL_RSS} and elaborated in Section~\ref{sec:safety_shield} and Section~\ref{sec:algoa_train_sim}.

We formulate the MARL problem as a tuple $G = (S, \mathcal{A}, P, r, \gamma)$, where $S$ denotes the joint state space. Each agent $i$ has state $s_i = \{ \vec{l}_i, \vec{v}_i, \vec{\alpha}_i, \vec{d}_i, \vec{c}_i \}$, representing position, velocity, acceleration, processed vision-based lane features, and LiDAR-based collision indicators. When communication is enabled, information shared from agent $j$ to agent $i$, denoted $s_{j,i}$, is incorporated as a \emph{delay-aware shared state} consistent with measured V2V latency, ensuring policies are trained on temporally misaligned neighbor information encountered during hardware deployment.

The discrete action space $\mathcal{A}$ aligns with real-world actuation capabilities and includes emergency stop, lane keeping, lane change maneuvers, and discrete acceleration or braking actions. Action selection is conditioned on delayed shared observations, requiring policies to reason under communication latency rather than assuming synchronous coordination. The unknown system transition dynamics are defined as $P : S \times \mathcal{A} \times S \rightarrow [0,1]$.

The reward is defined as
\begin{align}
    r(s, a) = w_1 \|\vec{v}_i\|_2 - w_2 \|\vec{c}_i\| + w_3 \|\vec{l}_i\| + r_i^{safe},
\end{align}
where $w_1$, $w_2$, and $w_3$ weight speed regulation, collision avoidance, and goal progress, respectively, and $r_i^{safe}$ penalizes Safety Shield interventions to promote intrinsically safe behavior under delayed and uncertain shared information.

During both training and execution, agents operate on delayed observations $s_i^{del} = \{ o_i, o_j^{del}, j \in \mathcal{N}_i \}$, where neighbor states are delayed according to measured fixed or time-varying V2V latency distributions. Embedding these communication characteristics directly into the training loop enables decentralized policies $\pi_{\theta}(a_i \mid s_i^{del})$ to coordinate robustly under realistic network constraints, supporting reliable zero-shot transfer to physical multi-vehicle platforms.

\subsection{Safety Shield and Control Integration}
\label{sec:safety_shield}

\begin{figure*}[!t]
    \centering
    {\includegraphics[width=1.15\linewidth]{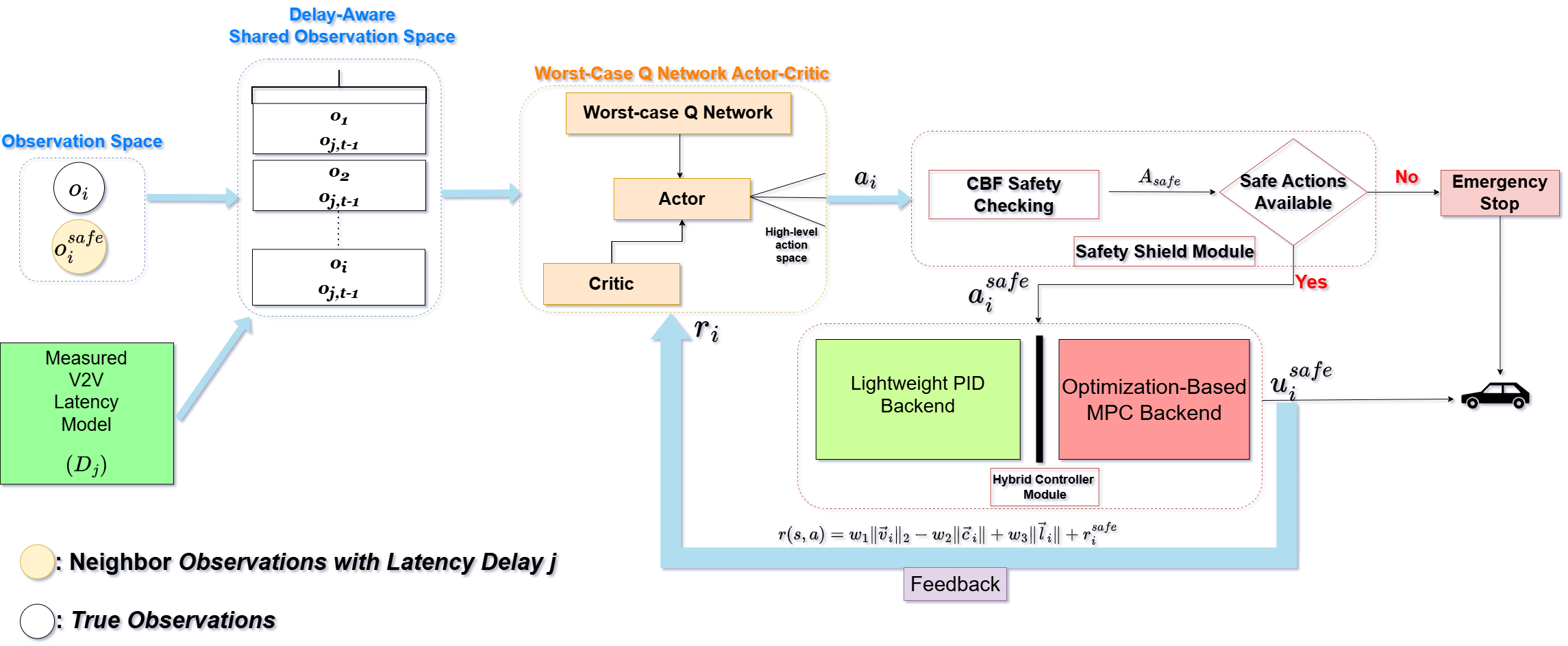}}\vspace{-6pt}
    \vspace*{0.1cm}
    \caption{Policy execution pipeline for agent $i$. During training, the critic and worst-case Q-network update the actor under observation delays. At test time, the actor selects a high-level action from delayed observations, which is filtered by a CBF-based Safety Shield before being passed to the Controller Module (PID or MPC) for low-level control $\boldsymbol{u}_i$. If no safe action exists, an emergency stop is triggered.}
    \label{fig:MARVEL_RSS}
    \vspace*{-0.5cm}
\end{figure*}

A central component of the framework is the \textbf{Safety Shield}, which guarantees that only feasible and provably safe control actions are executed. The Safety Shield leverages Control Barrier Functions (CBFs) enforced through the quadratic program (QP) in \eqref{CBF_QP} to filter unsafe commands prior to actuation~\cite{Zhang_2024, kong2015autonomous}. As shown in Fig.~\ref{fig:MARVEL_RSS}, high-level policy outputs are passed through the Safety Shield and subsequently executed by a selected low-level controller (PID or MPC).

\begin{align}
\min_{\vec{u} \in \mathbb{R}^m} \quad 
& \frac{1}{2} \| \vec{u} - \vec{u}^{\text{ref}} \|^2  \\
\text{s.t.} \quad 
& \frac{\partial h(\vec{x},t)}{\partial t} 
  + L_f h(\vec{x},t) 
  + L_g h(\vec{x},t)\vec{u} 
  \ge -\gamma h(\vec{x},t), \label{CBF_QP} \\
& v_{\min} \le v_{\text{ego}} \le v_{\max}, \nonumber \\
& a_{\min} \le \dot{v}_{\text{ego}} \le a_{\max}. \nonumber
\end{align}

The QP in \eqref{CBF_QP} minimally modifies the reference control input $\vec{u}^{\text{ref}}$ while enforcing forward invariance of the safe set defined by $h(\vec{x},t) \ge 0$. The CBF-QP is solved in a fully decentralized manner. Each agent computes its safety filter using locally measured ego states and communicated neighbor state estimates received via V2V links, along with known static obstacle information from the map. Neighboring vehicles are modeled as dynamic obstacles within the barrier constraints, while static obstacles are incorporated through predefined map coordinates. To account for communication latency and asynchronous updates, neighbor state estimates are treated as bounded-delay observations, and safety margins within $h(\vec{x},t)$ are conservatively inflated to preserve collision avoidance under state uncertainty. This decentralized and delay-aware formulation enforces safety at the agent level without centralized coordination, enabling scalable multi-agent deployment.

To address communication latency and estimation uncertainty, the Safety Shield incorporates conservative robustness mechanisms directly within the constraint of \eqref{CBF_QP}. Received neighbor states are treated as potentially stale due to measured communication delays and asynchronous updates. Safety margins within the CBF condition are therefore inflated to account for state uncertainty, effectively enlarging the admissible safe set considered by the QP. This delay-aware safety formulation preserves collision avoidance guarantees under realistic latency-constrained information sharing.

To bridge high-level policy outputs with low-level actuation, we employ two controller backends. A PID controller maps discrete policy actions to reference commands using proportional–integral–derivative feedback for lightweight real-time execution~\cite{Berducci2024}. Alternatively, an MPC controller solves a finite-horizon optimization problem at each timestep, incorporating CBF and CLF constraints to improve trajectory smoothness and foresight at increased computational cost~\cite{Sabouni2024}. Supporting both controllers highlights the modularity of the Safety Shield and its compatibility across heterogeneous control stacks.

\subsection{Algorithm and Training in Simulation}
\label{sec:algoa_train_sim}
Our complete algorithmic design is summarized in Algorithm~\ref{alg:rsr_rsmarl_pluggable}. We adopt the framework of centralized training with decentralized execution (CTDE); we train robust PPO agents where each agent is equipped with an extra worst-case Q network~\cite{liang2022efficient, Zhang_2024} estimating the expected return when agent's action-selection is potentially affected by inaccurate or perturbed observations. During training, agents experience delayed states to emulate network latency and improve robustness to real\text{-}world conditions. The pluggable low\text{-}level controllers ensure that the same MARL training loop can be used across different control architectures.

\begin{algorithm}[!t]
\caption{RSR\text{-}RSMARL with Communication-Aware Design and Pluggable Low\text{-}Level Controller}
\label{alg:rsr_rsmarl_pluggable}
\SetAlgoLined
\textit{Initialize} policy and critic networks\;
Load measured V2V latency model $\mathcal{D}_{lat}$; select delay mode $\in \{\text{none}, \text{fixed}, \text{time-varying}\}$\;
Select controller $\in \{\text{PID}, \text{MPC}\}$\;

\For{each episode $E$}{
    Initialize state $s=\prod_i s_i$, memory $M=\emptyset$\;
    
    \textbf{Rollout}: \For{each step, agent $i$}{
        Construct delay-aware state $s_i^{\text{del}} = \{o_i, o_j^{\text{del}}, j \in \mathcal{N}_i\}$ by applying measured latency model $\mathcal{D}_{lat}$ to received neighbor observations\;
        
        Choose action $a_i$ from policy $\pi_{\theta_i}(\cdot \mid s_i^{\text{del}})$\;
        
        \eIf{controller = PID}{
            Compute reference $\vec{u}^{\text{ref}}_i \leftarrow \text{PID}(s_i^{\text{del}}, a_i)$\;
            Filter with Safety Shield: $\boldsymbol{u}_i^{\textit{safe}} \leftarrow \text{CBF-QP}(\vec{u}^{\text{ref}}_i, s_i^{\text{del}})$\;
        }{
            Solve MPC optimization: $\boldsymbol{u}_i^{\textit{safe}} \leftarrow \text{MPC}(s_i^{\text{del}}, a_i)$\;
        }
        
        Execute $\boldsymbol{u}_i^{\textit{safe}}$, observe $s'$, compute reward $r_i$\;
        Store $(s_i^{\text{del}}, a_i, r_i, s'_i)$ in $M$; update $s \leftarrow s'$\;
    }
    
    \textbf{Training}: Update critics and policies using PPO and worst-case Q-learning objectives under communication-aware state inputs\;
}
\end{algorithm}

\subsection{Real\text{-}World Deployment}
For deployment, the trained policies are executed on a connected vehicle testbed using 1/10th-scale autonomous vehicles. Each vehicle estimates its state from onboard LiDAR, cameras, and IMU sensors, with additional information exchanged through V2V communication. The MARL policy produces high-level actions that are passed through the CBF-based Safety Shield, which filters unsafe commands before they reach the low-level controller. Depending on the configuration, safe actions are tracked by either a PID controller for lightweight execution or an MPC controller for smoother, optimization-based trajectories. This layered design ensures that real-world execution remains robust to sensing noise, communication delays, and actuation limits, while preserving the safety guarantees established in simulation. Further details on the hardware setup are provided in Section~\ref{sec:sim_and_hardware_setup}.

\section{Experiments and Results}

We evaluate the proposed framework along three dimensions: (i) sim-to-hardware robustness via zero-shot transfer from CARLA compared against domain randomization; (ii) Safety Shield effectiveness under controller-agnostic deployment across PID and MPC; and (iii) communication ablations including no-delay, fixed (F-2), and time-varying (TV) latency under matched and mismatched settings.

RSR-RSMARL denotes the proposed delay-aware framework with a CBF-based Safety Shield and V2V communication (PID or MPC control). Safe-RMM excludes delay modeling~\cite{Zhang2025}; RSR-MARL removes the Safety Shield; No-Comm disables V2V; and MARL-DR applies domain randomization~\cite{Tobin2017,Mehta2020}.

\subsection{Simulator and Hardware Testbed Setup}
\label{sec:sim_and_hardware_setup}
\textbf{Simulator:}
We conduct robust policy training and evaluation in the CARLA simulator environment~\cite{Dosovitskiy17}, where each vehicle is equipped with a suite of onboard sensors, including GPS and IMU modules. The simulation was conducted on a server configured with an AMD Ryzen 3970X processor and an NVIDIA Quadro RTX 6000 GPU. Experiments were performed with CARLA 0.9.15, Python 3.10, PyTorch 2.6, and CUDA 12.2.

\textbf{Hardware Testbed:}
For real-world evaluation, we deploy trained policies on a fleet of 1/10th-scale F1TENTH autonomous vehicles. Each vehicle is equipped with a 2D LiDAR (Hokuyo UST-10LX), an Intel RealSense D435i Depth Camera, and an onboard IMU, providing state estimates including position, velocity, and yaw rate. Vehicle-to-vehicle (V2V) communication is established through Wi-Fi at 5~Hz, with communication latency empirically measured between 10--20~ms.

Communication delays are measured using synchronized timestamps with full network stack instrumentation. While our controlled testbed achieves 10--20~ms latency, production V2X networks exhibit higher variability (50--200~ms) due to congestion, handoffs, and security overhead. To account for this gap, our training incorporates stochastic delay sampling from realistic delay distributions observed in urban V2V deployments.

The onboard computation platform is an NVIDIA Jetson Orin Nano running ROS Noetic on Ubuntu 20.04, executing both the MARL policy and the CBF-based Safety Shield in real time. Policy inference and safety filtering run at 10~Hz, while low-level control commands are issued to the VESC motor controller at 50~Hz. This configuration enables closed-loop execution under realistic sensing, communication, and control constraints, closely mirroring the conditions modeled in simulation.
\subsection{Hardware Evaluation}
\label{sec:real_world_evaluation}
\vspace*{-0.1cm}

\begin{figure}[h!]
    \centering
    \includegraphics[width=0.8\linewidth]{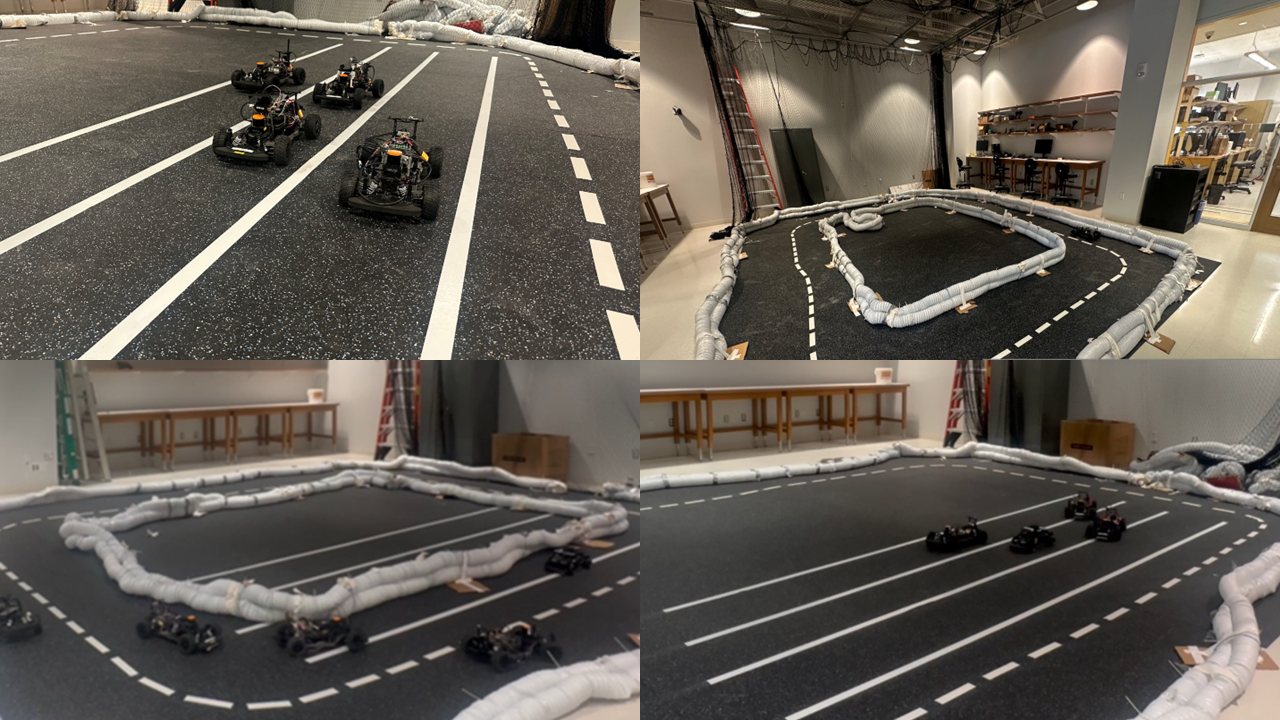}
    \caption{Hardware multi-agent experimental platform}
    \label{fig:hardware_multi_env}
\end{figure}

We evaluate the proposed framework on hardware platforms across two distinct driving scenarios: a 3-lane miniature highway and a 2-lane circular highway as shown in Figure~\ref{fig:hardware_multi_env}. Each environment is tested under three levels of complexity corresponding to the number of obstacles introduced. During hardware evaluation, obstacles may be repositioned between trials to demonstrate robustness to environmental changes while maintaining repeatable experimental conditions.

Table~\ref{tab:Eval_MARVEL_RealWorld} presents the evaluation results across all scenarios. On the 3-lane highway, both delay-aware RSR-RSMARL variants—time-varying (TV) and fixed-delay (F-2)—achieve zero collisions across all obstacle levels, with the TV model consistently yielding the lowest completion times. In contrast, RSR-MARL, which excludes the Safety Shield module, exhibits a growing number of collisions as obstacle density increases. Although RSR-MARL completes some scenarios faster, this speed comes at the cost of erratic and risk-prone behavior. These findings underscore that faster completion does not imply a safer driving strategy. The slightly slower yet collision-free performance of the TV and F-2 variants reflects a deliberate trade-off prioritizing safety under delayed communication.

The No-Comm RSR-RSMARL variant, which disables V2V communication, results in longer completion times and occasional collisions under perturbations, further highlighting the robustness gained from explicitly modeling communication delay. 

Domain randomization (MARL-DR) exhibits a more pronounced degradation in hardware performance, with collisions increasing as obstacle density grows. In contrast, both delay-aware variants (TV and F-2) remain collision-free, with the TV model consistently outperforming the fixed-delay formulation, demonstrating the benefit of time-varying latency modeling during training.

A comparable trend appears in the 2-lane oval highway scenario. Across both PID and MPC control, the TV and F-2 RSR-RSMARL variants maintain zero collisions, with the TV model again achieving the most efficient completion times. By contrast, the RSR-MARL baseline degrades as obstacle complexity increases, and the No-Comm variant struggles once the environment becomes more challenging.  

The integration of \textbf{MPC} within RSR-RSMARL further enhances trajectory smoothness and robustness. Although \textbf{MPC introduces a modest increase in computational load relative to PID}, the delay-aware TV and F-2 policies consistently generate safer and more stable actions. 

Taken together, these results demonstrate that incorporating communication-delay modeling during training directly improves hardware performance under realistic latency conditions. In particular, the TV formulation consistently provides the strongest coordination efficiency while preserving zero-collision safety, indicating that structured latency modeling enhances reliable sim-to-hardware transfer in multi-agent environments.

\begin{table}[h!]
    \centering
    \renewcommand{\arraystretch}{1.35}
    \caption{Evaluation Results on 3-Lane Highway and 2-Lane Oval Highway}
    \begin{tabularx}{0.96\columnwidth}{Xccc}
        \hline
        & \multicolumn{3}{c}{Number of Obstacles} \\ \cline{2-4}
        Method & \textbf{None} & \textbf{1} & \textbf{2} \\ 
        \hline
\multicolumn{4}{c}{\textit{3-Lane Highway}} \\ \hline
\textbf{RSR-RSMARL (TV, MPC)}  & \textbf{0}, \textbf{33.1} & \textbf{0}, \textbf{34.0} & \textbf{0}, \textbf{34.5} \\
RSR-RSMARL (F-2, MPC)          & 0, 33.6 & 0, 34.8 & 0, 35.2 \\
\textbf{RSR-RSMARL (TV, PID)}  & \textbf{0}, \textbf{34.2} & \textbf{0}, \textbf{35.4} & \textbf{0}, \textbf{36.1} \\
RSR-RSMARL (F-2, PID)          & 0, 34.8 & 0, 36.0 & 0, 36.8 \\
Safe-RMM (MPC)                 & 0, 34.5 & 1, 35.3 & 2, 36.1 \\
RSR-MARL (PID)                 & 1, 36.8 & 2, 37.5 & 3, 38.0 \\
No-Comm RSR-RSMARL (PID)       & 0, 36.7 & 1, 37.9 & 1, 39.5 \\
MARL-DR                        & 2, 36.5 & 3, 37.8 & 5, 39.2 \\
\hline
\multicolumn{4}{c}{\textit{2-Lane Oval Highway}} \\ \hline
\textbf{RSR-RSMARL (TV, MPC)}  & \textbf{0}, \textbf{47.5} & \textbf{0}, \textbf{48.6} & \textbf{0}, \textbf{50.2} \\
RSR-RSMARL (F-2, MPC)          & 0, 48.2 & 0, 49.7 & 0, 51.4 \\
\textbf{RSR-RSMARL (TV, PID)}  & \textbf{0}, \textbf{48.1} & \textbf{0}, \textbf{49.4} & \textbf{0}, \textbf{51.6} \\
RSR-RSMARL (F-2, PID)          & 0, 48.9 & 0, 50.3 & 0, 52.4 \\
Safe-RMM (MPC)                 & 0, 50.2 & 1, 51.5 & 2, 53.0 \\
RSR-MARL (PID)                 & 1, 51.3 & 2, 52.5 & 3, 53.1 \\
No-Comm RSR-RSMARL (PID)       & 0, 54.8 & 1, 56.2 & 1, 57.1 \\
MARL-DR                        & 2, 52.0 & 3, 57.6 & 5, 59.0 \\
\hline
    \end{tabularx}
    \begin{tablenotes}[flushleft]
        \footnotesize
        \item \emph{\textbf{Summary}—The time-varying (TV) delay RSR-RSMARL consistently achieves the strongest safety and efficiency across both MPC and PID controllers. The fixed-delay (F-2) variant remains robust but exhibits slightly increased completion times, highlighting the benefit of time-varying delay modeling.}

    \end{tablenotes}
    \label{tab:Eval_MARVEL_RealWorld}
\end{table}

\subsection{Ablation and Benchmark Evaluation}




\begin{figure*}[!t]
    \centering

    \begin{subfigure}[t]{1.0\linewidth}
        \centering
        \includegraphics[width=\linewidth]{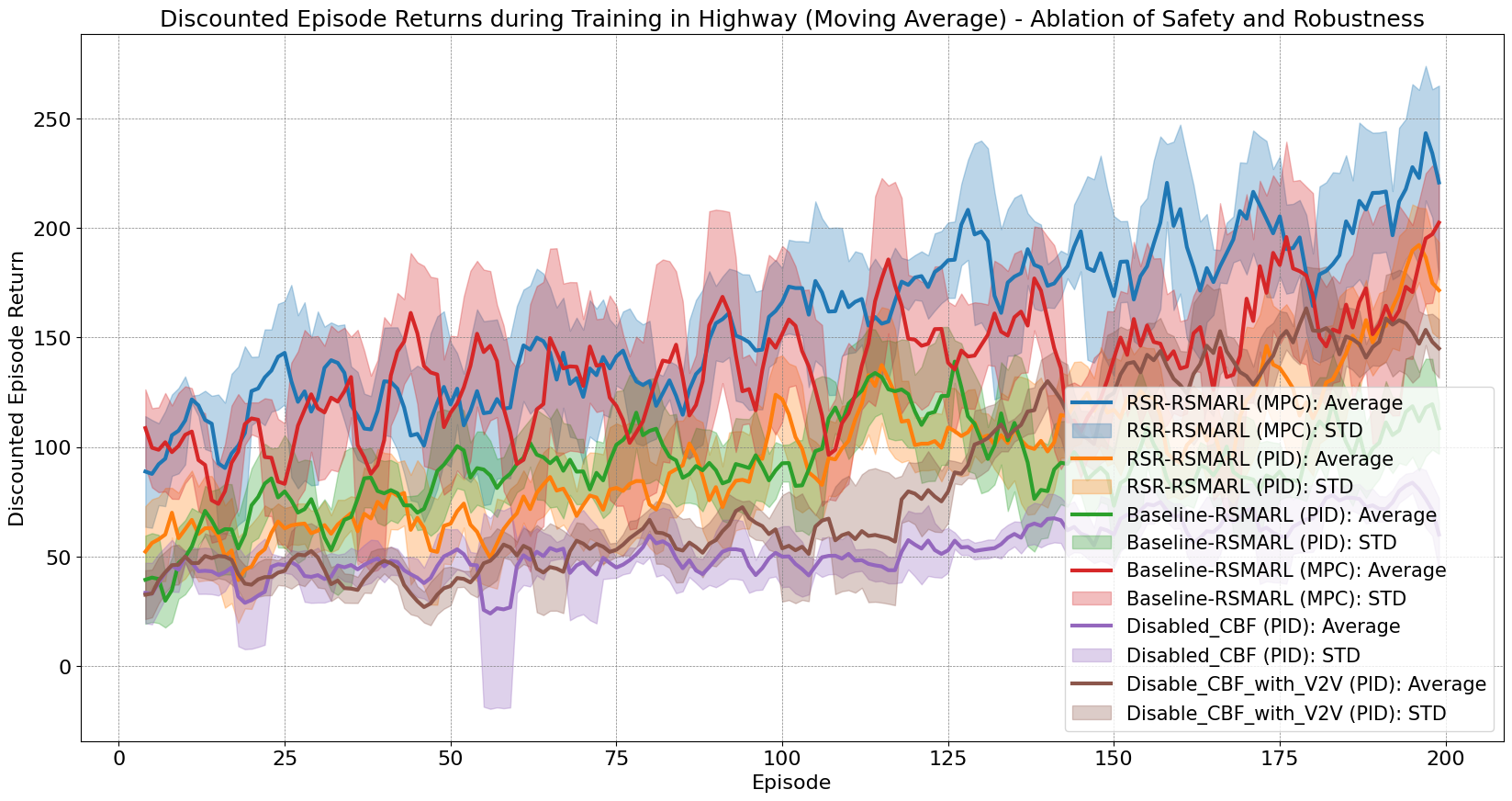}
        \caption{Ablation of Safety and Communication Mechanisms During Training}
        \label{fig:ablation_sub}
    \end{subfigure}
    \hfill
    \begin{subfigure}[t]{1.0\linewidth}
        \centering
        \includegraphics[width=\linewidth]{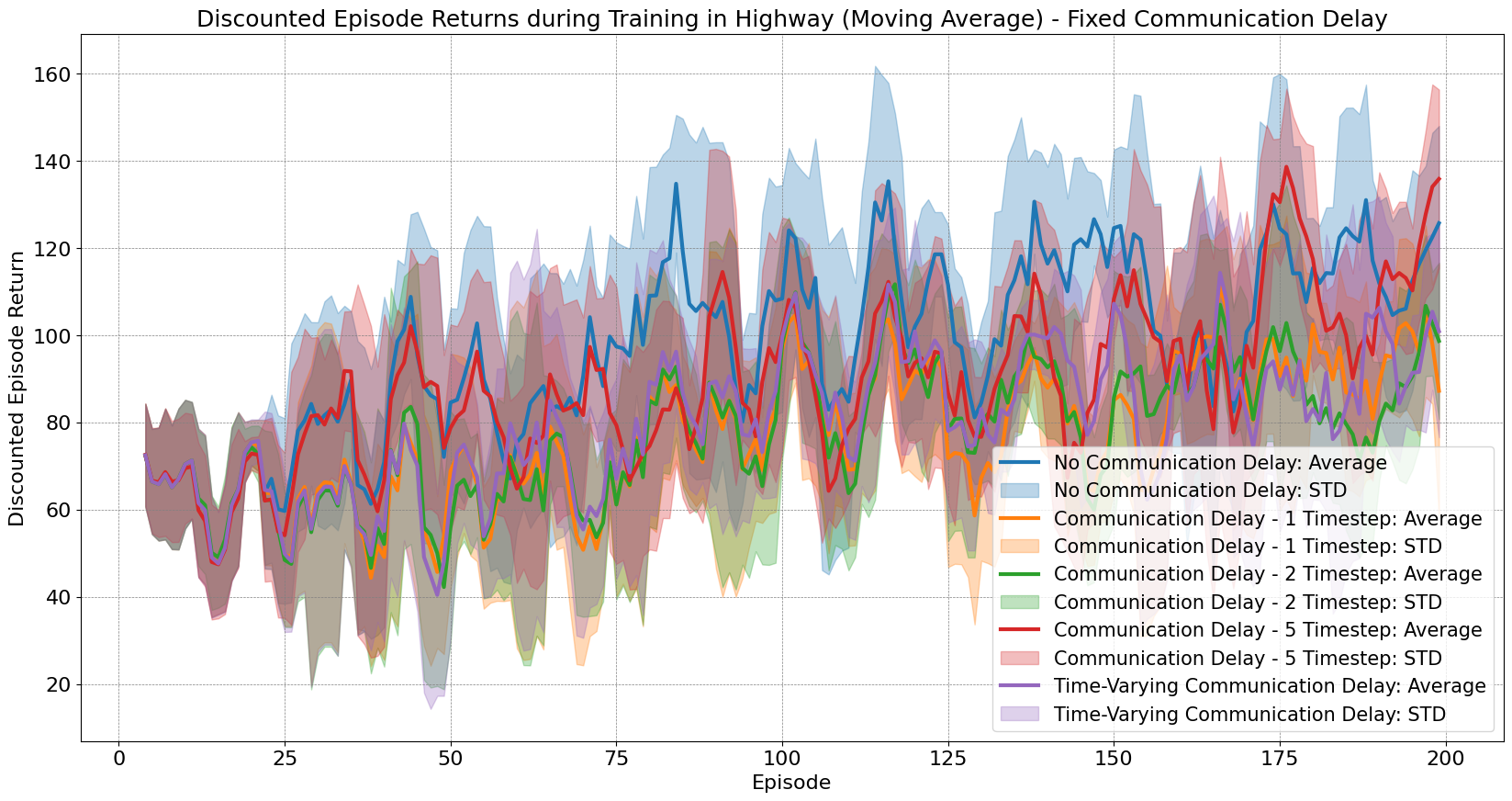}
        \caption{Varying Communication Delay}
        \label{fig:Delay_sub}
    \end{subfigure}

    \caption{
    Discounted episode returns during training in the highway scenario.
    (a) Ablation of safety and robustness under the adaptive low-level controller.
    (b) Robustness under increasing varying communication delays.
    }
    \label{fig:Sim_Training_Results_All}
\end{figure*}

We evaluate our framework in a simulated highway scenario through targeted ablation and benchmarking studies isolating the effects of (i) controller backend selection (MPC vs.\ PID), (ii) removal of the CBF-based Safety Shield, (iii) increasing fixed communication delay, and (iv) comparison with domain randomization (MARL-DR). These experiments quantify the impact of architectural and robustness design choices on training stability, task efficiency, and collision avoidance.


Figures~\ref{fig:ablation_sub} compares training performance under MPC and PID control. In both cases, RSR-RSMARL achieves higher discounted efficiency returns and smoother convergence than ablated variants. Disabling the Safety Shield results in unstable learning dynamics and degraded long-term returns, underscoring the importance of formally enforced safety during training.

Figure~\ref{fig:Delay_sub} evaluates robustness under increasing fixed communication delays and compares them against the TV formulation. While higher fixed delays introduce performance degradation and increased variance during training, RSR-RSMARL maintains stable learning dynamics and converges to competitive returns across all delay levels. Notably, the TV model exhibits improved stability relative to large fixed delays, highlighting the benefit of time-varying latency modeling.

Table~\ref{tab:collision_efficiency_results} reports safety (collisions) and efficiency (discounted return) under different communication delay models. Under fixed-delay settings (F-1–F-5), \textbf{RSR-RSMARL} achieves strong task performance, with F-5 yielding the highest peak return (161.90) while maintaining near-zero collisions. These policies are optimized for a specific latency regime and therefore maximize efficiency when communication characteristics are known and stationary. In contrast, the time-varying (TV) delay model, where delay is sampled in $[0,5]$, achieves zero collisions while maintaining competitive efficiency (139.71 for MPC, 124.18 for PID). Although slightly lower than the best fixed-delay case, TV training optimizes expected performance over a distribution of delays rather than a single latency value, producing coordination strategies that are inherently robust to stochastic communication jitter encountered in real-world deployment.

Removing the Safety Shield substantially increases collision frequency (e.g., \textbf{RSR-MARL} and \textbf{Non-Robust RSR-MARL}), demonstrating that delay modeling alone is insufficient to guarantee safety. We further benchmark against domain randomization (\textbf{MARL-DR}), which injects stochastic observation noise to improve sim-to-real robustness. While MARL-DR mitigates degradation relative to non-robust baselines, it exhibits higher collision rates and lower efficiency than \textbf{RSR-RSMARL}. This comparison underscores a key distinction: noise-based robustness does not explicitly address communication latency nor enforce safety constraints. By jointly modeling delay during training and integrating a CBF-based Safety Shield, \textbf{RSR-RSMARL} achieves a superior safety--efficiency trade-off under realistic communication conditions.

\begin{table}[t]
    \centering
    \renewcommand{\arraystretch}{1.2}
    \caption{Evaluation Results Under Communication Delay Modeling}
    \begin{threeparttable}
    \begin{tabular}{lccc}
        \hline
        \textbf{Method} & \textbf{Delay} & \textbf{Coll.} & \textbf{Return} \\
        \hline
        \textbf{RSR-RSMARL (MPC)} 
            & F-1 & 0 & 120.23 \\
            & F-2 & 2 & 125,18 \\
            & F-5 & 1 & 161.90 \\
            & TV (0--5) & 0 & 139.71 \\
        \hline
        \textbf{RSR-RSMARL (PID)} 
            & F-2 & 2 & 129.52 \\
            & TV (0--5) & 0 & 124.18 \\
        \hline
        \textbf{MARL-DR} & F-2 & 10 & 95.07 \\
        \textbf{Safe-RMM (MPC)} & F-2 & 15 & 75.65 \\
        \textbf{RSR-MARL} & F-2 & 42 & 28.84 \\
        \textbf{Non-Robust RSR-MARL} & F-2 & 45 & 25.26 \\
        \hline
    \end{tabular}
    \begin{tablenotes}[flushleft]
        \footnotesize
        \item F-$k$: fixed delay of $k$ steps. TV: time-varying delay sampled in $[0,5]$.
        \item Results averaged over 50 test episodes.
    \end{tablenotes}
    \end{threeparttable}
    \label{tab:collision_efficiency_results}
\end{table}

\begin{figure}[h!]
    \centering
    \includegraphics[width=0.8\linewidth]{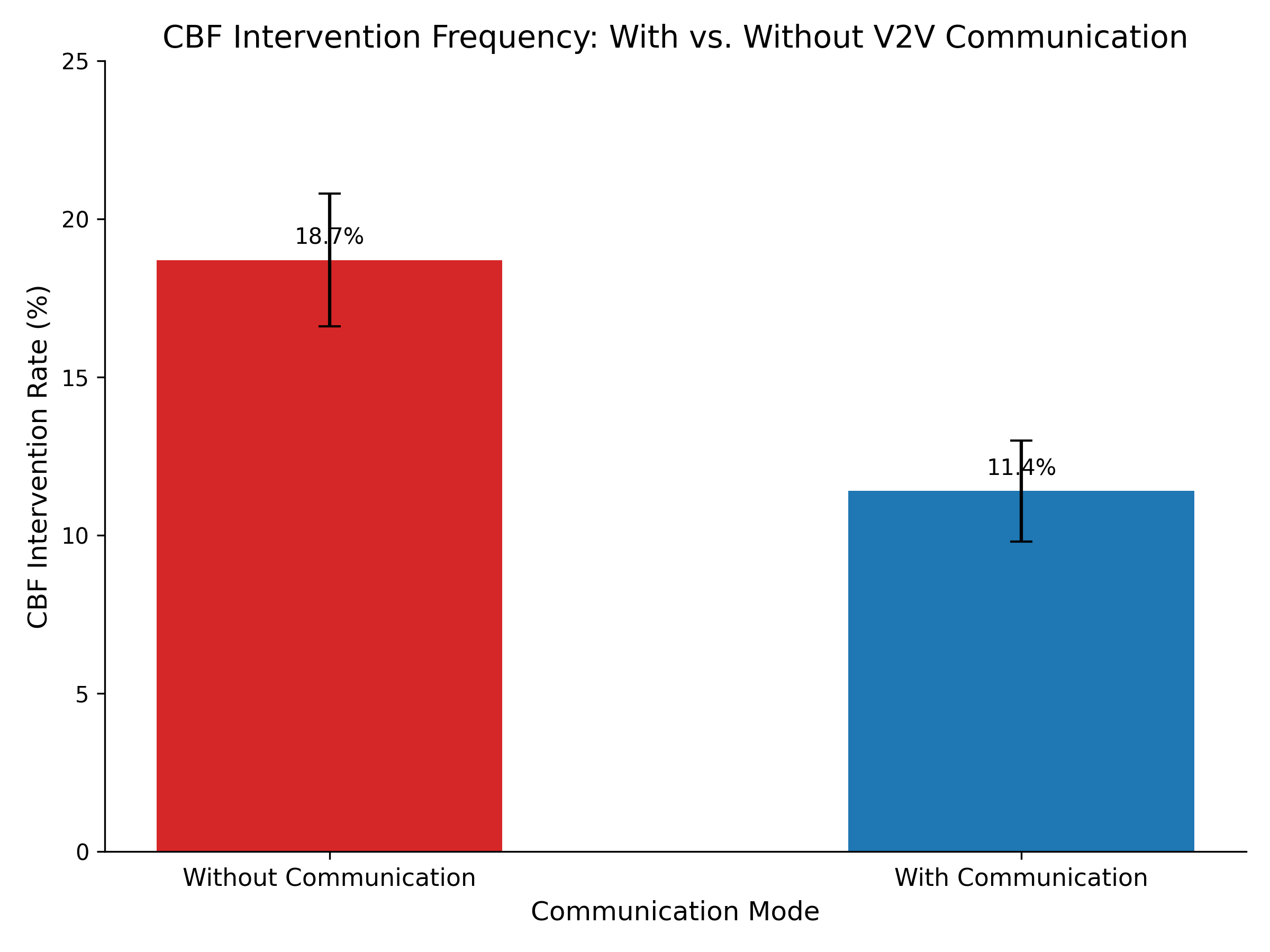}
    \caption{CBF Intervention Frequency: With vs. Without V2V Communication}
    \label{fig:comm_vs_no_comm}
\end{figure}

\subsection{Impact of Communication on Safety and Coordination}
We evaluate V2V communication in complex driving scenarios where agents must account for neighboring vehicle positions and velocities during maneuvers. Incorporating neighbor state information is essential for safe action selection, particularly in the presence of dynamic NCAVs that require real-time adaptation and coordinated behavior. Communication latency is modeled during training (Sec.~\ref{sec:algoa_train_sim}) by injecting stochastic delays into inter-agent messaging, and a time-varying delay model is deployed on hardware to reflect realistic wireless latency. Policies trained with delay variability remain stable under real-world latency fluctuations, even when deployment delays differ from those encountered in simulation.

Figure~\ref{fig:comm_vs_no_comm} demonstrates that shared information reduces both the need for reactive safety overrides and the likelihood of unsafe maneuvers. In dense multi-agent interactions, coordination enabled by V2V stabilizes policy outputs and lowers the average CBF intervention rate from 18.7\% (no communication) to 11.4\% (with communication), highlighting the role of communication in supporting safe and efficient real-world operation.

\section{Conclusion}


This paper presents \textbf{RSR-RSMARL}, a communication-aware multi-agent reinforcement learning framework designed for Real-Sim-Real transfer of connected autonomous vehicles. By explicitly incorporating measured V2V communication latency into the MARL state formulation and embedding both fixed and time-varying delay models into simulation training, the proposed approach learns policies that remain robust under asynchronous and delayed information exchange on physical hardware. Experimental validation in CARLA and on 1/10th-scale autonomous vehicle platforms demonstrates reliable zero-shot sim-to-hardware transfer, improved coordination stability, and consistent safety performance under realistic communication constraints. These results underscore the importance of hardware-grounded latency modeling and safety-aware training for scalable and dependable multi-agent autonomous systems.









\clearpage
\bibliography{references}

\end{document}